\showboxdepth=\maxdimen
\showboxbreadth=\maxdimen

\documentclass[11pt,twoside]{article}

\usepackage{asp2014}

\aspSuppressVolSlug
\resetcounters

\begin{document}

\title{Building astroBERT, a language model for Astronomy \& Astrophysics}

\author{Felix~Grezes$^1$,
        Sergi~Blanco-Cuaresma,
        Alberto~Accomazzi,
        Michael~J.~Kurtz,
        Golnaz~Shapurian,
        Edwin~Henneken,
        Carolyn~S.~Grant,
        Donna~M.~Thompson,
        Roman~Chyla,
        Stephen~McDonald,
        Timothy~W.~Hostetler,
        Matthew~R.~Templeton,
        Kelly~E.~Lockhart,
        Nemanja~Martinovic,
        Shinyi~Chen,
        Chris~Tanner,
        and Pavlos~Protopapas.
} 
\affil{$^1$Harvard-Smithsonian Center for Astrophysics, Cambridge, MA, USA; \email{felix.grezes@cfa.harvard.edu}}

\paperauthor{Felix Grezes}{felix.grezes@cfa.harvard.edu}{0000-0001-8714-7774}{Harvard-Smithsonian Center for Astrophysics}{HEAD}{Cambridge}{MA}{02138}{USA}
\paperauthor{Golnaz~Shapurian}{gshapurian@cfa.harvard.edu}{0000-0001-9759-9811}{Harvard-Smithsonian Center for Astrophysics}{HEAD}{Cambridge}{MA}{02138}{USA}
\paperauthor{Sergi~Blanco-Cuaresma}{sblancocuaresma@cfa.harvard.edu}{0000-0002-1584-0171}{Harvard-Smithsonian Center for Astrophysics}{HEAD}{Cambridge}{MA}{02138}{USA}
\paperauthor{Alberto~Accomazzi}{aaccomazzi@cfa.harvard.edu}{0000-0002-4110-3511}{Harvard-Smithsonian Center for Astrophysics}{HEAD}{Cambridge}{MA}{02138}{USA}
\paperauthor{Michael~J.~Kurtz}{kurtz@cfa.harvard.edu}{0000-0002-6949-0090}{Harvard-Smithsonian Center for Astrophysics}{HEAD}{Cambridge}{MA}{02138}{USA}
\paperauthor{Edwin~A.~Henneken}{ehenneken@cfa.harvard.edu}{0000-0003-4264-2450}{Harvard-Smithsonian Center for Astrophysics}{HEAD}{Cambridge}{MA}{02138}{USA}
\paperauthor{Carolyn~S.~Grant}{cgrant@cfa.harvard.edu}{0000-0003-4424-7366}{Harvard-Smithsonian Center for Astrophysics}{HEAD}{Cambridge}{MA}{02138}{USA}
\paperauthor{Donna~M.~Thompson}{dthompson@cfa.harvard.edu}{0000-0001-6870-2365}{Harvard-Smithsonian Center for Astrophysics}{HEAD}{Cambridge}{MA}{02138}{USA}
\paperauthor{Roman~Chyla}{rchyla@cfa.harvard.edu}{0000-0003-3041-2092}{Harvard-Smithsonian Center for Astrophysics}{HEAD}{Cambridge}{MA}{02138}{USA}
\paperauthor{Stephen~McDonald}{stephen.mcdonald@cfa.harvard.edu}{0000-0003-1270-0605}{Harvard-Smithsonian Center for Astrophysics}{HEAD}{Cambridge}{MA}{02138}{USA}
\paperauthor{Timothy~W.~Hostetler}{thostetler@cfa.harvard.edu}{0000-0001-9238-3667}{Harvard-Smithsonian Center for Astrophysics}{HEAD}{Cambridge}{MA}{02138}{USA}
\paperauthor{Matthew~R.~Templeton}{matthew.templeton@cfa.harvard.edu}{0000-0003-1918-0622}{Harvard-Smithsonian Center for Astrophysics}{HEAD}{Cambridge}{MA}{02138}{USA}
\paperauthor{Kelly~E.~Lockhart}{kelly.lockhart@cfa.harvard.edu}{0000-0002-8130-1440}{Harvard-Smithsonian Center for Astrophysics}{HEAD}{Cambridge}{MA}{02138}{USA}
\paperauthor{Nemanja~Martinovic}{nemanja.martinovic@cfa.harvard.edu}{0000-0002-9485-7296}{Harvard-Smithsonian Center for Astrophysics}{HEAD}{Cambridge}{MA}{02138}{USA}
\paperauthor{Shinyi~Chen}{shinyi.chen@cfa.harvard.edu}{0000-0002-7641-7051}{Harvard-Smithsonian Center for Astrophysics}{HEAD}{Cambridge}{MA}{02138}{USA}
\paperauthor{Chris~Tanner}{christanner@g.harvard.edu}{0000-0002-0141-9466}{Harvard}{IACS}{Cambridge}{MA}{02138}{USA}
\paperauthor{Pavlos~Protopapas}{pprotopapas@g.harvard.edu}{0000-0002-8178-8463}{Harvard}{SEAS}{Cambridge}{MA}{02138}{USA}


\begin{abstract}
The existing search tools for exploring the NASA Astrophysics Data System (ADS) can be quite rich and empowering (e.g., similar and trending operators), but researchers are not yet allowed to fully leverage semantic search. 
For example, a query for "results from the Planck mission" should be able to distinguish between all the various meanings of Planck (person, mission, constant, institutions and more) without further clarification from the user.
At ADS, we are applying modern machine learning and natural language processing techniques to our dataset of recent astronomy publications to train astroBERT, a deeply contextual language model based on research at Google.
Using astroBERT, we aim to enrich the ADS dataset and improve its discoverability, and in particular we are developing our own named entity recognition tool. We present here our preliminary results and lessons learned.
\end{abstract}


\section{Introduction}
The NASA ADS (\url{https://ui.adsabs.harvard.edu}) has been publicly servicing the astrophysical research community since \citet{1993ASPC...52..132K}. It maintains a searchable bibliographic collection of more than 15 million records. The database can be explored via complex queries using filters (e.g. authors, year range, text in title and/or body, refereed or not) combined with Boolean operators to explore the citations and references, and more advanced operators such as paper similarity, co-reads, and reviews.

As powerful as the existing tools are, they do not make use of recent advances in natural language processing and semantic search. Ideally, if a user wanted to filter by papers that use data from the ESA Planck space observatory, a simple query of "Planck mission" should suffice, without requiring manual input of specific keywords such as \texttt{year:2009-2022} or \texttt{NOT author:"Planck, Max"}. Such queries highlight the inherent ambiguity of natural language and how remarkably flexible humans are at understanding semantics from contextual clues. Currently, ADS handles many of the most common ambiguities with carefully crafted rules (e.g. using SIMBAD objects).

To automate the process of identifying, disambiguating, and tagging named entities within the ADS database, we are building \textbf{astroBERT}, a language model for astronomical content. This astroBERT is based on Google's BERT deep neural network transformer architecture \citep{2018arXiv181004805D}, inspired by the AllenAI's SciBERT \citep{2019arXiv190310676B} efforts, built using tools from Huggingface \citep{2019arXiv191003771W}, and is trained is trained using a large collection of recent astronomy ADS papers. 

Our preliminary results show that astroBERT outperforms BERT on the named entity recognition task on ADS data. Based on this encouraging result, we can now work on improving and integrating astroBERT into ADS services, and providing access to it for the astrophysics community.

\section{Related Work: BERT and SciBERT}
\citet{2018arXiv181004805D} introduced the Bidirectional Encoder Representations from Transformers (BERT) language model, based on the Transformer technology \citep{vaswani2017attention}. It was trained with text from Wikipedia and the Brown Corpus, and then fine-tuned to achieve state-of-the-art performance on 11 NLP tasks, including sentiment analysis, semantic role labeling, sentence classification and word disambiguation.

Building off the success of BERT, \citet{2019arXiv190310676B} created SciBERT,  Focusing on scientific papers from the Semantic Scholar corpus, SciBERT is part of a growing list of models that adapt BERT to specific domains and tasks. Its  state-of-the-art performance on a wide range of scientific domain NLP tasks motivates our work on astroBERT.
The SciBERT paper shows that pre-training BERT with domain-specific language data improves its performance when compared to the original BERT.

\section{Technical Details}

The work presented here was implemented using the open-source python-based  Huggingface \citep{2019arXiv191003771W} library, which provides easy access to state-of-the-art NLP models, and tools.
Runtimes are reported as run on a machine with two NVIDIA V100 GPUs, two 12-core Intel Xeon Gold 6246 CPUs, and 768GB of RAM.
Unless specified, all  parameters were left to Huggingface v4.10.2 defaults .

\subsection{Data Preparation}
In order to train our language model, our first task was to build a large dataset of clean English language text from astronomy papers. From the ADS database, we selected papers for which the source was provided by the publishers in XML format. The final dataset is made up of \textbf{395,499} documents (16GB on disk). Per best practices, this was evenly split into  training, evaluation and testing datasets.
Despite our efforts to select the best quality data sources, we required to execute a clean-up to remove unprintable unicode characters and rare unicode characters (those appearing 50 times or less). We also replaced accented characters by the non-accented counterparts when possible. In future re-iterations we will consider a more exhaustive cleaning to filter out meaningless strings representing data tables, mathematical formulas, and appendices of raw data.


\subsection{Tokenizer}
An important part of the text processing pipeline is the tokenizer, which converts text data into inputs compatible with the model To stick as close as possible to the original BERT design choices. 
we trained a WordPiece 
tokenizer maximum vocabulary size of 30,000 tokens. Due to the extensive use of accronyms in the astrophysics literature, and its tendency to correspond to English words (e.g., AURA, BLAST, CLUSTER, INTEGRAL, WISE), we decided to keep the case information in our model.
the tokenizer did not automatically convert every character to the same case, however accents were removed to simplify downstream tasks.
Training the tokenizer over the entire data took approximately \textbf{45 minutes}.
The astroBERT vocabulary overlaps with that of BERT and SciBERT by \textbf{24.5\%} and \textbf{35.3\%} respectively (over total number of tokens), while BERT and SciBERT overlapped by 27.0\%. The total size of the data corresponds to \textbf{3,819,322,591} tokens, or 2,977,635,680 words and 121,207,934 sentences when parsed by NLTK \citep{bird2009natural} .

\subsection{The MLM, NSP and NER Tasks}
Following the methodology of the BERT paper, astroBERT is pre-trained on two tasks that do not require hand labeled data: first the masked language model (MLM) task in which the model predicts tokens that were randomly masked by using the contextual text; and second MLM jointly with the next sentence prediction (NSP) task in which the model predicts if a sentence is subsequent to another in the original document. The original random masking and shuffling probabilities were kept unchanged  (imitating the original BERT setup). 

Following astroBERT pre-training, we transferred the models to our first downstream task: named entity recognition (NER), in which the model tries to identify organizations. Using a list of 908 astronomy organization acronyms and their full names, we scanned the acknowledgement sections of 44K astronomy manuscripts and built a dataset of 1856 sentences with 6279 annotated organizations. The models were evaluated using 10-fold cross-validation. 


\section{Training astroBERT and Preliminary Results}
We trained astroBERT(MLM) for 15 epochs on the MLM task, and astroBERT(NSP
+MLM) for 3 epochs on the joint NSP and MLM tasks. astroBERT(MLM) was initialized with BERT weights, and astroBERT(NSP+MLM) with the trained astroBERT
(MLM).
Training astroBERT takes approximately \textbf{8 hours} per epoch for the MLM task, and \textbf{22 hours} for the joint NSP+MLM task.  
Training was done using mixed floating point precision, which has been shown to reduce memory consumption and lead to faster learning \citep{2017arXiv171003740M}.
The NER models were fine-tuned for 3 epochs, each epoch taking approximately \textbf{90 seconds}. 

In table \ref{perps} we report perplexity scores our astroBERT, BERT, and SciBERT models on our cleaned ADS dataset, as well across various standard English texts dataset provided by Huggingface. Because these models use different vocabularies, scores should be compared within model and across datasets, and not across models.
These scores show that BERT, which is trained on Wikipedia English, can be re-trained into astroBERT to performs best on ADS text. SciBERT, which was trained on data from Semantic Scholar (not tested here) performs similarly on all 3 scientific text datasets. 
As expected, adding the NSP task to astroBERT(NSP+MLM) training causes it to perform worse in terms of perplexity on the MLM task. 

However as shown in table \ref{ner}, in which we report the average f1-scores and standard deviations across cross-validation runs for BERT and astroBERT after fine-tuning, astroBERT(NSP+MLM) slightly outperforms astroBERT(MLM) on the NER task, and both outperform the original BERT model.

\begin{table}[!ht]
\caption{Comparisons of perplexity scores across BERT models}
\label{perps}
\smallskip
\begin{center}
{\small
\begin{tabular}{lcccc}  
\tableline
\noalign{\smallskip}
    & BERT & SciBERT & astroBERT & astroBERT \\
    &       &       &   (MLM)   & (NSP+MLM)\\
\noalign{\smallskip}
\tableline
\noalign{\smallskip}
ADS                         & 38.4 & 4.05 & \textbf{4.16}   & \textbf{5.71} \\
wikitext-2-raw-v1           & 29.7 & 8.64 & 18.5            & 25.9 \\
ptb\_text\_only             & 82.8 & 9.63 & 20.0            & 27.4 \\
scientific\_papers (arxiv)  & 41.3 & 4.23 & 4.82            & 6.87 \\
scientific\_papers (pubmed) & 34.0 & 3.96 & 8.41            & 12.6 \\
\noalign{\smallskip}
\tableline\
\end{tabular}
}
\end{center}
\end{table}

\begin{table}[!ht]
\caption{Comparisons of NER scores across BERT models on ADS data}
\label{ner}
\smallskip
\begin{center}
{\small
\begin{tabular}{lccc}  
\tableline
\noalign{\smallskip}
    & BERT & astroBERT & astroBERT\\
    &       & (MLM) & (NSP+MLM)\\ 
\noalign{\smallskip}
\tableline
\noalign{\smallskip}
f1-score & 0.859 & \textbf{0.893} & \textbf{0.902}\\
standard deviation & 0.014 & 0.028 & 0.014\\
\noalign{\smallskip}
\tableline\
\end{tabular}
}
\end{center}
\end{table}

\section{Future Work}

With astroBERT showing encouraging results on the preliminary ADS data, we are currently building a more complete NER dataset covering more named entities and papers. Our goal is to extend our services to automatically identify entities such as facilities, and allow ADS users to unambiguously find papers that mention them. 

Beyond improving ADS services, astroBERT will be made publicly available to the astrophysics community through the Huggingface Model Hub
, which provides free access to many state-of-the-art language models. 


\bibliography{O1-001}  


\end{document}